\documentclass[conference]{IEEEtran}
\IEEEoverridecommandlockouts
\usepackage{cite}
\usepackage{amsmath,amssymb,amsfonts}
\usepackage{algorithmic}
\usepackage{graphicx}
\usepackage{textcomp}
\usepackage{subfigure}
\usepackage{xcolor}
\usepackage{bbm}
\def\BibTeX{{\rm B\kern-.05em{\sc i\kern-.025em b}\kern-.08em
    T\kern-.1667em\lower.7ex\hbox{E}\kern-.125emX}}
\begin{document}

\title{Enhancing Personalized Ranking With Differentiable Group AUC Optimization\\
}

\author{
%\IEEEauthorblockN{Anonymous}

%\IEEEauthorblockN{1\textsuperscript{st} Xiao Sun}
%\IEEEauthorblockA{\textit{Meituan Inc.} \\
%Beijing, China \\
%xsun@pku.edu.cn}
%\and
%\IEEEauthorblockN{2\textsuperscript{nd} Bo Zhang}
%\IEEEauthorblockA{\textit{Meituan Inc.} \\
%Beijing, China \\
%zhangbo58@meituan.com}
%\and
%\IEEEauthorblockN{3\textsuperscript{rd} Chenrui Zhang}
%\IEEEauthorblockA{\textit{Meituan Inc.} \\
%Beijing, China \\
%zhangchenrui02@meituan.com}
%\and
%\IEEEauthorblockN{4\textsuperscript{th} Han Ren}
%\IEEEauthorblockA{\textit{Meituan Inc.} \\
%Beijing, China \\
%renhan@meituan.com}
%\and
%\IEEEauthorblockN{5\textsuperscript{th} Mingchen Cai}
%\IEEEauthorblockA{\textit{Meituan Inc.} \\
%Beijing, China \\
%caimingchen@meituan.com}

\IEEEauthorblockN{Xiao Sun, Bo Zhang, Chenrui Zhang, Han Ren, Mingchen Cai}
\IEEEauthorblockA{\textit{Meituan Inc.} \\
Beijing, China \\
\{sunxiao10, zhangbo58, zhangchenrui02, renhan, caimingchen\}@meituan.com}

}

\maketitle

\begin{abstract}
AUC is a common metric for evaluating the performance of a classifier. However, most classifiers are trained with cross entropy, and it does not optimize the AUC metric directly, which leaves a gap between the training and evaluation stage. 
In this paper, we propose the PDAOM loss, a Personalized and Differentiable AUC Optimization method with Maximum violation, which can be directly applied when training a binary classifier and optimized with gradient-based methods. 
Specifically, we construct the pairwise exponential loss with difficult pair of positive and negative samples within sub-batches grouped by user ID, aiming to guide the classifier to pay attention to the relation between hard-distinguished pairs of opposite samples from the perspective of independent users. Compared to the origin form of pairwise exponential loss, the proposed PDAOM loss not only improves the AUC and GAUC metrics in the offline evaluation, but also reduces the computation complexity of the training objective. Furthermore, online evaluation of the PDAOM loss on the 'Guess What You Like' feed recommendation application in Meituan manifests $1.40\%$ increase in click count and $0.65\%$ increase in order count compared to the baseline model, which is a significant improvement in this well-developed online life service recommendation system.
\end{abstract}


\section{Introduction}
\begin{figure}[t]
\centering
\includegraphics[width=3.1in]{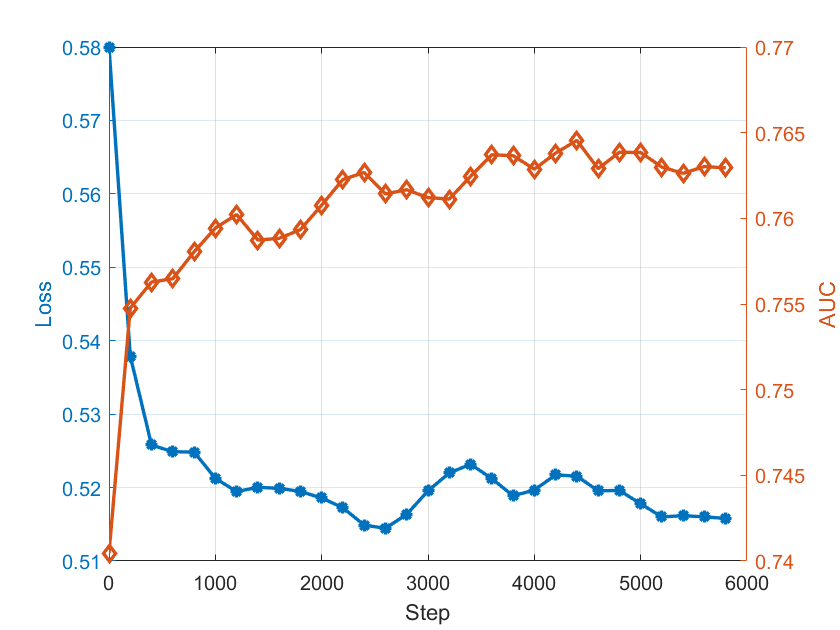}
\caption{The curves of the loss and the AUC metric varying with training steps, where the AUC metric does not always increase with the loss decreasing.}
\label{fig_1}
\end{figure}

Bipartite ranking has drawn a lot of attentions in the academic community in the past few decades, and has been widely employed in the industrial applications. It aims at learning a model that is capable of ranking positive samples higher than negative ones. Without loss of generality, we take the recommendation system for example and detail how bipartite ranking is applied. According to the statistics of the historical behaviors of a user, the recommendation system provides an ordered list of items, in which the preferred items will appear ahead the uninterested ones. The key idea to achieve this goal is to predict the click-through rate (CTR) for every item. The items that the user viewed but not clicked are labeled as negative ones, while the clicked items are labeled as positive ones. Then the CTR prediction model can be trained as a binary classifier and optimized with the cross entropy. In this way, each sample is treated independently, and the restrictive relation between positive and negative samples is not introduced during training. Another concern is that compared to the items user viewed, the clicked items only occupy a small fraction. Thus, the performance of the model is basically evaluated with the Area Under Curve (AUC) metric in the situations where the data distribution is imbalanced. AUC measures the probability for a randomly sampled positive instance to have a higher score than a randomly drawn negative instance. Hence, the cross entropy objective in the training phase does not completely match the target in the evaluation phase. In practice, one common observation is that the AUC metric does not increase with the training loss decreasing, and an example of training on the industrial recommendation dataset is shown in Fig. \ref{fig_1}. It inspires us to optimize the AUC metric directly during training.

\begin{figure}[t]
\centering
\includegraphics[width=3.3in]{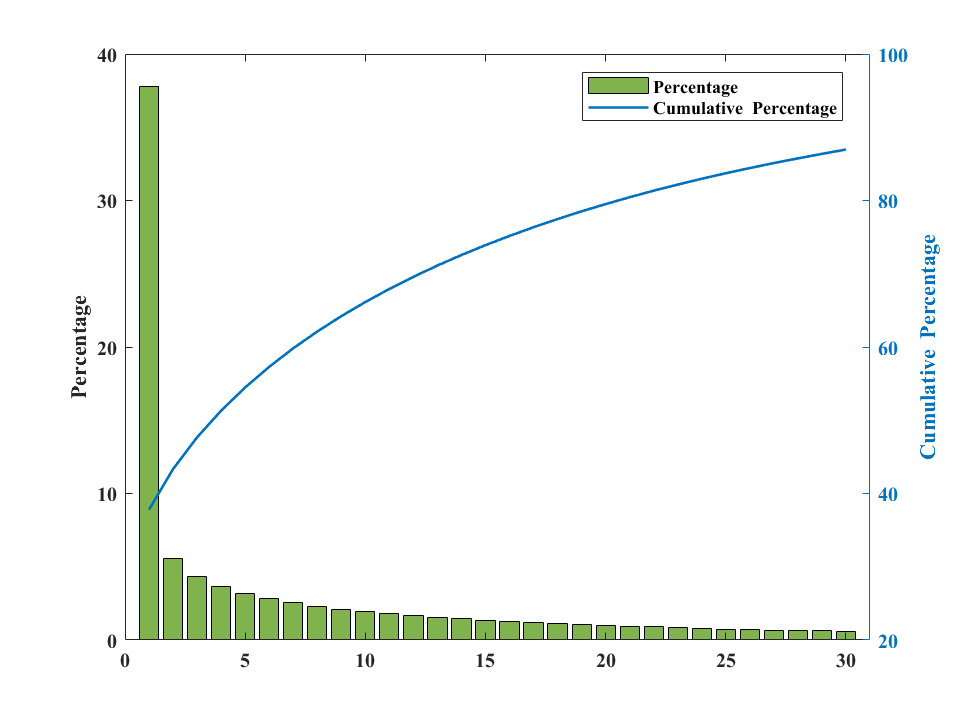}
\caption{Long tail phenomenon in Meituan E-commerce business. The products in the top bins contribute a large proportion of orders.}
\label{fig_longtail}
\end{figure}

Another big problem that the recommendation systems often face is 'long tail' phenomenon, i.e., A small fraction of products occupies a large proportion of sales. Fig. \ref{fig_longtail} shows the statistics from Meituan E-commerce business. We divide the products into 100 bins according to the quantity of their orders and plot the top $30$ bins. We can see that the products from the top $1$ bin contributes about $37\%$ orders, and the products from the top $20$ bins contributes about $80\%$ orders. If we train a CTR prediction model with such unbalanced data, the model tends to assign higher scores to hot products, even though some users may not like these items, which degrades the model considerably in terms of making personalized predictions. Group AUC (GAUC) \cite{zhu2017optimized} is a proper metric to evaluate the capacity of personalized recommendation of a ranking model. It is derived by calculating the AUC within the sets grouped by user ID and averaging the results of each set. Empirically, the offline GAUC metric is more consistent with the online effect compared to the AUC metric, which further motivates us to add the GAUC metric to the objective when training the ranking model. 

However, there are two main reasons that prevent us from optimizing the AUC metric directly. On the one hand, the AUC metric is calculated by a summation of binary indicator functions, which is not differentiable. Thus, the gradient-based optimization methods can not be applied to the problem. On the other hand, the formula of AUC considers every pair of positive and negative samples, leading to the time complexity up to $\mathcal{O}(N^+N^-)$, which is unacceptable in the real industrial scenarios where the quantity of the positive or negative samples are usually on the order of billions. Extensive efforts have been made to solve the questions mentioned above. For the former, recent works attempt to substitute the differentiable surrogate objective function for the origin AUC formulation. Zhao et al. \cite{zhao2011online} propose to replace the indicator function with its convex surrogate, i.e., the hinge loss function. Gao et al. \cite{gao2013one} design a regression-based algorithm with pairwise squared objective function, which measures the ranking errors between two instances from different classes. \cite{gao2015consistency}  studies the AUC consistency based on optimizing pairwise surrogate objective functions. For the latter, mini-batch optimization strategy \cite{gultekin2020mba} makes it possible to deal with large-scale dataset. In order to apply the AUC maximization method into data-intensive scenarios, we dive into optimizing it in the mini-batch manner. Specifically, we propose the PDAOM loss, a Personalized and Differentiable AUC Optimization method with Maximum violation, which focuses on distinguishing difficult pair of opposite samples instead of taking the whole combinations into consideration. This trick not only improves the offline performance, but also decreases the complexity of optimization.

\section{RELATED WORK }
The AUC metric is widely used when measuring the capacity of a classifier in dealing with imbalanced data distribution. It represents the probability that a randomly picked positive instance is ranked higher than a randomly picked negative one, and is not sensitive to the proportion of positive and negative samples. In the past few years, many efforts have been made to optimize AUC directly \cite{yuan2021large,lei2021stochastic,natole2018stochastic}. The main idea behind this is to substitute a convex surrogate objective function for AUC, such as pairwise logistic loss \cite{gao2015consistency}, pairwise hinge loss \cite{zhao2011online,khalid2016confidence}, pairwise squared loss \cite{gao2013one,natole2018stochastic,liu2018fast,gultekin2020mba} and pairwise exponential loss \cite{freund2003efficient,rudin2009margin}. Most of the studies above are designed for linear models, and may not be applicable for deep learning without modifications. \cite{sulam2017maximizing} adopts an online buffered gradient method \cite{zhao2011online} and solves the problem of breast cancer classification in deep learning means. In each iteration, the positive and negative instances need to be stored in the buffer for computing the approximate value of AUC, thus such algorithm can not be generalized to large-scale dataset. \cite{liu2019stochastic} reformulates the pairwise square loss into a saddle-point problem, making it possible for dealing with big data. However, these works treat every pair of positive and negative samples equally, which leads the model to spend unnecessary capacity on easily-distinguished pairs. In this work, we construct the difficult pair of samples within a batch of size $N$, aiming at enhancing the relation that positive samples score higher than negative samples. Besides, this trick reduces the complexity from $\mathcal{O}(N^2)$ to $\mathcal{O}(N)$. 

\begin{table}[t]
\newcommand{\tabincell}[2]{\begin{tabular}{@{}#1@{}}#2\end{tabular}}
\centering
  %\caption{Comparison of different classification methods evaluated on ModelNet40.}
\begin{tabular}{l|l} 
%\toprule
\hline
Surrogate   &formula    \\
%\midrule
\hline
Pairwise Logistic Loss & $\phi_{PLL}(t)=\log(1+\exp(-t))$            \\

Pairwise Hinge Loss  & $\phi_{PHL}(t)=max(0,1-t)$         \\

Pairwise Squared Loss  & $\phi_{PSL}(t)=(1-t)^2$        \\

Pairwise Exponential Loss. & $\phi_{PEL}(t)=exp(-t)$       \\

%\bottomrule
\hline
  \end{tabular}
  \vspace{0.1cm}
  \caption{Commonly-used surrogate function for the indicator of AUC.}
  \label{table_1}
%  \vspace{-0.8cm}
\end{table}

\section{METHODOLOGY}

\subsection{Preliminaries} \label{Preliminaries}
The origin definition of AUC is related to the receiver operating characteristic (ROC) curve. Assuming that the classifier produces a continuous value that represents the probability of the input sample being positive, then a decision threshold is needed to determine if the sample should be labeled as positive or negative. For each threshold, we can obtain a pair of true-positive rate and false-positive rate. By traversing the threshold over the range from 0 to 1 and plotting the obtained pairs of rates, the ROC curve is produced. Thus, it is complicated to derive the AUC metric by means of calculating the area under the ROC curve. An equivalent form of AUC is formulated as the normalized Wilcoxon-Mann-Whitney (WMW) statistic:
\begin{equation}
	AUC=\frac{\sum_{i=0}^{m-1}\sum_{j=0}^{n-1}\mathbbm{1}(f(x_i^+)>f(x_j^-))}{mn},
\end{equation}
where $\{x_i^+\}$ and $\{x_j^-\}$ are sets of positive and negative samples, respectively. Since the indicator function is non-differentiable, the WMW statistic can not be optimized with gradient-based algorithms. It motivates us to look for a differentiable surrogate objective function that substitutes for the WMW statistic. That is to say, obtaining the optimal solution of the surrogate function is equivalent to maximizing the AUC metric. We formulate the objective function with surrogate as
\begin{equation}
	\mathop\mathbb E_{\substack{x^+\sim \mathcal P^+\\x^-\sim \mathcal P^-}}(\phi(f(x^+)-f(x^-)),
\end{equation}
where $\phi$ is the surrogate function, and some commonly-used examples are listed in Table \ref{table_1}. $\mathcal P^+$ and $\mathcal P^-$ are the distributions of positive and negative samples, respectively. In this paper, we employ the pairwise exponential loss as surrogate for two reasons. Firstly, \cite{gao2015consistency} has proved that the pairwise exponential loss is consistent with AUC. Secondly, we compare the listed surrogates in our preliminary experiments and find that the pairwise exponential loss outperforms others in the offline evaluations (see Table \ref{table_preliminary_exp1}).

\subsection{AUC Optimization with Maximum Violation}
There are two main drawbacks for the mentioned surrogate objective function. On the one hand, such objective pays equal attention to every pair, thus the classifier spends a lot of efforts on modeling the relation of easily-distinguished positive and negative sample pairs. On the other hand, for a batch data which contains $N^+$ positive samples and $N^-$ negative samples, the complexity of dealing with every pair is $\mathcal{O}(N^+N^-)$, which is time-consuming for a large batch size. Aiming at the questions above, we propose to construct the difficult sample pairs and lead the model to focus on difficult sample pairs instead of all pairs. A difficult sample pair refers to the instances with opposite labels that the model is not confident to distinguish which one is positive or negative. Thus the output scores for such opposite samples are close, which makes the decision boundary indiscriminate. Consider that
\begin{equation}
	\mathop\mathbb E_{\substack{x^+\sim \mathcal P^+\\x^-\sim \mathcal P^-}} (\phi(f(x^+)-f(x^-)) \leq \max_{\substack{x^+\sim \mathcal P^+\\x^-\sim \mathcal P^-}} (\phi(f(x^+)-f(x^-)),
\end{equation}
a feasible solution is to set $\max_{\substack{x^+\sim \mathcal P^+\\x^-\sim \mathcal P^-}} (\phi(f(x^+)-f(x^-))$ as objective function. Instead of incorporating every pair of positive and negative samples into deriving the expectation, the calculation of the maximum depends on only one pair of opposite samples that most likely violates the relation. In this way, the accumulation of the loss from easy negatives does not influence the update of the model. 
Despite that such conversion leads the model to focus on establishing the decision boundary, the complexity remains $\mathcal{O}(N^+N^-)$. Since $f(x^+)-f(x^-) \in [-1,1]$, the surrogate function $\phi$ decreases monotonically in this interval. Equivalently, $\max_{\substack{x^+\sim \mathcal P^+\\x^-\sim \mathcal P^-}} (\phi(f(x^+)-f(x^-))$ is simplified as
\begin{equation} \label{eq4}
	\phi(\min_{\substack{x^+\sim \mathcal P^+\\x^-\sim \mathcal P^-}} (f(x^+)-f(x^-)))=\phi(\min_{x^+\sim \mathcal P^+}f(x^+)-\max_{x^-\sim \mathcal P^-}f(x^-)).
\end{equation}
Ideally, the lowest score of a positive sample is desired to be higher than the highest score of a negative sample within a batch. We define the DAOM loss as
\begin{equation} \label{daom_loss}
	L_{DAOM} =\phi(\min_{x^+\sim \mathcal P^+}f(x^+)-\max_{x^-\sim \mathcal P^-}f(x^-)).
\end{equation}
Note that we only pick the positive sample with the highest score and the negative sample with the lowest score, and the complexity of constructing the pair reduces to $\mathcal{O}(N^++N^-)$.

\subsection{Enhancing Personalized Ranking via GAUC Optimization}
The above section details how to construct paired samples within a batch, which do not meet the requirement for personalized recommendation. In practice, we find that the GAUC \cite{zhu2017optimized} metric is more consistent with the online effect. Consequently, it is a natural idea to add the GAUC metric to the objective when optimizing the model. Considering the original calculation of the GAUC metric, the samples are firstly divided into several groups. In this context, the groups are partitioned by the user ID. Then the AUC metric is calculated within each group respectively, and the GAUC metric is derived by averaging the AUC metrics from all groups with weight. The weight is optionally proportional to the impression or click times, here we set the weight to be 1 for all users. We simulate the calculation of GAUC in the training phase. When preparing the training data, we rank the samples according to the user ID of each sample, so that the samples of a user may appear in the same batch. The data of one batch may contain several different user IDs, we split the batch into sub-batches in which the user ID in a sub-batch is same. Then we apply the DAOM loss on each sub-batch and define the personalized DAOM loss as
\begin{equation} \label{pdaom_loss}
	L_{PDAOM}=\sum_{u\in U}\phi(\min_{x^+\sim \mathcal P^+_u}f(x^+)-\max_{x^-\sim \mathcal P^-_u}f(x^-)),
\end{equation}
where $U$ denotes the sub-batches grouped by user ID.
In the condition of training a binary classifier, the proposed PDAOM loss may accompany with the cross entropy to form the final objective function
\begin{equation} 
\begin{aligned}
\label{newLoss}
%	L=L_{CE}+ \\ \lambda L_{DAOM},
	L=-y\log(f(x))-(1-y)\log (1-f(x)) \\
	+ \lambda \sum_{u\in U}\phi(\min_{x^+\sim \mathcal P^+_u}f(x^+)-\max_{x^-\sim \mathcal P^-_u}f(x^-)),
\end{aligned}
\end{equation}
where $y$ is the label, and $\lambda$ balances the weights of the cross entropy and PDAOM loss.

\begin{table*}[ht]
\newcommand{\tabincell}[2]{\begin{tabular}{@{}#1@{}}#2\end{tabular}}
\centering
\begin{tabular}{c|cccc|cccc} 

\hline
Objective   &\multicolumn{4}{c|}{Cross Entropy}   &\multicolumn{4}{c}{Cross Entropy + PDAOM}    \\
\hline
Task &\multicolumn{2}{c}{CTR} &\multicolumn{2}{c|}{CVR} &\multicolumn{2}{c}{CTR} &\multicolumn{2}{c}{CVR} \\
\hline
Model  &AUC  &GAUC   &AUC & GAUC &AUC  &GAUC   &AUC & GAUC  \\

\hline
DeepFM\cite{guodeepfm} &75.09 &67.41 &88.67 &75.10 &75.47 &67.81 &88.86 &75.55\\
MMOE\cite{ma2018modeling}   &75.82 &68.12 &89.01 &75.70 &75.97 &68.53 &89.20 &76.07\\
PLE\cite{tang2020progressive}  &75.90 &68.25 &89.09 &75.94 &76.10 &68.69 &89.35 &76.33\\
Cross-Stitch\cite{misra2016cross} &75.62 &68.05 &88.97 &75.73 &75.88 &68.47 &89.18 &76.09\\
CGC\cite{tang2020progressive}  &75.89 &68.19 &89.15 &75.85 &76.11 &68.62 &89.34 &76.29\\
\hline
  \end{tabular}
  \vspace{0.1cm}
  \caption{The performance of training with different network architectures on Meituan recommendation dataset.}
  \label{table_networks_MT}
\end{table*}

\begin{table*}[ht]
\newcommand{\tabincell}[2]{\begin{tabular}{@{}#1@{}}#2\end{tabular}}
\centering
  %\caption{Comparison of different classification methods evaluated on ModelNet40.}
\begin{tabular}{c|cccc|cccc} 

\hline
Objective   &\multicolumn{4}{c|}{Cross Entropy}   &\multicolumn{4}{c}{Cross Entropy + PDAOM}    \\
\hline
Task &\multicolumn{2}{c}{CTR} &\multicolumn{2}{c|}{CVR} &\multicolumn{2}{c}{CTR} &\multicolumn{2}{c}{CVR} \\
\hline
Model  &AUC  &GAUC   &AUC & GAUC &AUC  &GAUC   &AUC & GAUC  \\

\hline
DeepFM\cite{guodeepfm}            &60.72 &60.11 &67.35 &62.31 &60.99 &60.49 &67.57  &62.67\\
MMOE\cite{ma2018modeling}         &60.90 &60.23 &67.48 &62.40 &61.12 &60.57 &67.68  &62.72\\
PLE\cite{tang2020progressive}     &61.10 &60.38 &67.62 &62.48 &61.30 &60.66 &67.83  &62.76\\
Cross-Stitch\cite{misra2016cross} &60.78 &60.18 &67.40 &62.35 &61.04 &60.54 &67.57  &62.70\\
CGC\cite{tang2020progressive}     &61.06 &60.40 &67.60 &62.49 &61.28 &60.71 &67.82  &62.75\\
\hline
  \end{tabular}
  \vspace{0.1cm}
  \caption{The performance of training with different network architectures on Ali-CCP recommendation dataset.}
  \label{exp_aliccp}
\end{table*}

\section{EXPERIMENTS}
In this section, we validate the effectiveness of the proposed PDAOM loss. Without loss of generality, we choose the recommendation scenario and evaluate the contribution via click-through rate (CTR) and conversion rate (CVR) prediction tasks. We conduct both offline and online experiments to verify how DAOM loss improves the performance compared with the baseline settings.

\textbf{Dataset.} 
The experiments are conducted based on two datasets, Meituan recommendation dataset and Ali-CCP dataset \cite{ma2018entire}. Meituan recommendation dataset is constructed by sampling from the user's exposure and behavior logs of the 'Guess What You Like' stand on the first page of Meituan app. It contains about 2.8 billion samples which cover 267 million users and 43 million items. The records of users' behaviors include viewing, clicking and ordering, thus we can perform click-through rate prediction and conversion rate prediction tasks on the dataset. Ali-CCP is a public dataset that is commonly used to evaluate the performance of a multi-task model for CTR and CVR prediction. It contains 84 million samples from the recommendation system of Taobao.

\textbf{Metric.}
For both CTR and CVR prediction tasks, we adopt the AUC metric to evaluate the performance of the model, whose meaning has been explained in Section \ref{Preliminaries}. Note that the AUC metric only measures the overall ability of the model to rank positive instances ahead the negative ones among the whole samples, it is not good at reflecting the situation in a more personalized aspect. Thus, we further employ the GAUC metric to assess the capacity of personalized recommendation.

%For calculating GAUC, the samples are firstly divided into several groups, then the AUC results are derived within each group. In this context, the groups are partitioned by the user ID. Finally, the AUC results in each group are averaged with weight, which is proportional to the impression or click times.

\subsection{Evaluation on Different Model}
To evaluate the universal effectiveness of the proposed PDAOM loss, experiments are conducted based on various networks on Meituan recommendation dataset and Ali-CCP dataset, respectively. 
We select several mainstream architectures for multi-task learning of CTR and CVR prediction that are widely used in the real recommendation systems, which are listed as below:
\begin{itemize}
\item
    \textbf{DeepFM} \cite{guodeepfm} consists of a FM layer and a hidden layer, which accepts the same raw feature vectors input and learns low-order and high-order feature interactions simultaneously. We use DeepFM as a backbone in multi-task learning.
\item
	\textbf{Cross-Stitch} \cite{misra2016cross} proposes a sharing unit that can learn an optimal combination of shared and task-specific representations.
\item 
	\textbf{MMOE} \cite{ma2018modeling} adapts the Mixture-of-Experts structure to multi-task learning by sharing the expert subnets across all tasks, with a gating network trained to optimize each task.
\item
	\textbf{CGC} \cite{tang2020progressive} explicitly separates shared and task-specific experts and proposes a customized gate to combine two kinds of experts.
\item
	\textbf{PLE} \cite{tang2020progressive} is an extension to CGC model with multi-level gating networks and progressive separation routing for more efficient information sharing and joint learning.
\end{itemize}

We calculate the AUC and GAUC metric of the CTR and CVR models which are trained with traditional Cross Entropy and the new Objective (Eq. \ref{newLoss}). The results on Meituan recommendation dataset are shown in Table \ref{table_networks_MT}. For such large-scale dataset, A/A test is conducted and the performance fluctuation of the CTR and CVR prediction tasks is $\pm 0.02$.  
As we can see from Table \ref{table_networks_MT}, both the AUC and the GAUC metrics of two tasks improves with the PDAOM loss as extra objective. Take the CTR prediction task for example, the new objective brings up to 0.38 AUC gains and 0.40 GAUC gains for DeepFM. 
From DeepFM and Cross-Stitch network to MMOE, the architecture of the models becomes complex gradually, and the proposed PDAOM loss keeps boosting the performance of each network.
Even for the state-of-the-art network, CGC, the AUC gains and the GAUC gains reaches up to 0.22 and 0.43, respectively, which is difficult to acquire considering the magnitude of the dataset. 
Based on the observation above, we can prove the effectiveness of the PDAOM objective for different network architectures.
Similar conclusions can be drawn on Ali-CCP dataset.

\subsection{Online A/B Test}
Online A/B test is conducted in the 'Guess What You Like' stand which lies on the first page of Meituan app. We add the proposed PDAOM loss to the objective function of the baseline model, and train the new model with one year's data before putting it online. The final ranking score is derived by a combination of the predicted CTR and CVR score. We observe the online results for a week: the total click count increases $1.40\%$, and the total order count increases $0.65\%$. Note that the P value is 0.008, which implies that the PDAOM loss brings a significant improvement in the well-developed recommendation system.

\begin{table}[t]
\newcommand{\tabincell}[2]{\begin{tabular}{@{}#1@{}}#2\end{tabular}}
\centering
  %\caption{Comparison of different classification methods evaluated on ModelNet40.}
\begin{tabular}{c|cc|cc} 

\hline
Task   &\multicolumn{2}{c|}{CTR}   &\multicolumn{2}{c}{CVR}    \\
\hline
Objective  &AUC  &GAUC   &AUC & GAUC   \\

\hline
CE & 75.89    & 68.19    & 89.15    & 75.85\\
CE+PSL   & 75.89    & 68.21    & 89.12  & 75.88 \\
CE+PLL  & 75.90    & 68.18    & 89.11  & 75.77 \\
CE+PEL  & 75.90    & 68.21    & 89.15  & 	75.89 \\
\hline
  \end{tabular}
  \vspace{0.1cm}
  \caption{Comparison between commonly-used surrogates on Meituan recommendation dataset.}
  \label{table_preliminary_exp1}
\end{table}

\begin{table}[t]
\newcommand{\tabincell}[2]{\begin{tabular}{@{}#1@{}}#2\end{tabular}}
\centering
  %\caption{Comparison of different classification methods evaluated on ModelNet40.}
\begin{tabular}{c|cc|cc} % 

\hline
Task   &\multicolumn{2}{c|}{CTR}   &\multicolumn{2}{c}{CVR}    \\
\hline
Objective  &AUC  &GAUC   &AUC & GAUC   \\

\hline
CE+PEL(Origin)  & 75.90    & 68.21    & 89.15  & 	75.89 \\
%CE+PEL(Eq.\ref{eq4})   & 75.94    & 68.27    & 89.18  & 76.07 \\
CE+DAOM  & 76.03     & 68.35    & 89.26  & 76.13 \\
CE+PDAOM  & 76.11     & 68.62    & 89.34  & 76.29 \\
\hline
  \end{tabular}
  \vspace{0.1cm}
  \caption{The performance of training with variations of PEL objective on Meituan recommendation dataset.}
  \label{table_preliminary_exp2}
%  \vspace{-0.5cm}
\end{table}

\begin{figure}[t]
\centering
\includegraphics[width=3in]{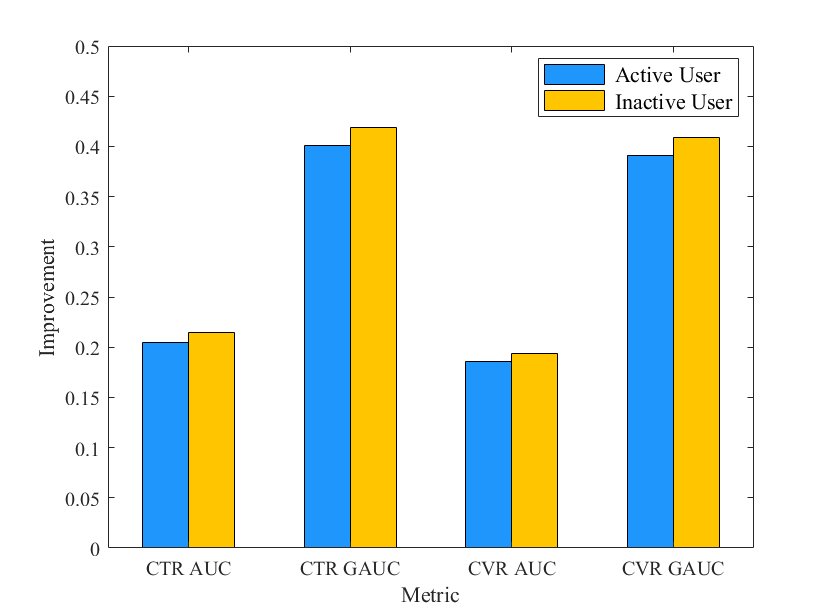}
\caption{The improvements of the PDAOM loss on active users and inactive users.}
\label{fig_gains}
\end{figure}

\subsection{Improvement on Different Groups of User}
In order to explore the influence of the proposed PDAOM loss on different users, we rank the users according to the amounts of behaviors. All users are divided into two groups evenly. The users with more behaviors are assigned into the first group, while the users with less behaviors are assigned into the second one. The gains on the two groups are shown in Fig. \ref{fig_gains}. The PDAOM loss is effective for both active users and inactive users, and improves more for the latter.

\subsection{Comparisons Between Variant Surrogates of AUC}
We compare several commonly-used surrogate objective functions based on the CGC backbone, and the results are shown in Table \ref{table_preliminary_exp1}. Despite the fact that the pairwise exponential loss (PEL) achieves comparable performance or slightly outperforms other surrogates, we have to say that the forms of pairwise losses have slight influence on the ranking results. 
Based on pairwise exponential loss, we explore how the way of constructing of positive and negative sample pairs affects the performance, and the results are shown in Table \ref{table_preliminary_exp2}. 
The DAOM loss gains better performance compared to the origin PEL, which proves the effectiveness of the way we construct opposite samples. 
The PDAOM loss continues to improve, especially in terms of the GAUC metric. Organizing the samples by grouping the same user ID together and calculating the DAOM loss within a group when deriving the objective make it possible to incorporate the GAUC goal in the training phase.

\begin{figure}[h]
\centering
\subfigure[The effect of batch size on the performance.]{
  \includegraphics[height=2.2in]{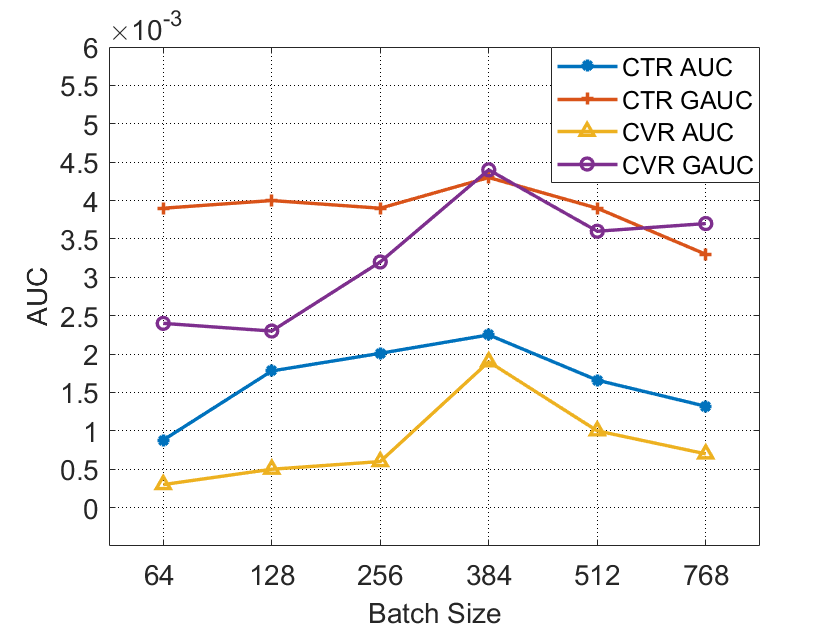}
  \label{fig_1_a}
  }
\centering
\subfigure[The effect of the weight on the performance.]{
  \includegraphics[height=2.2in]{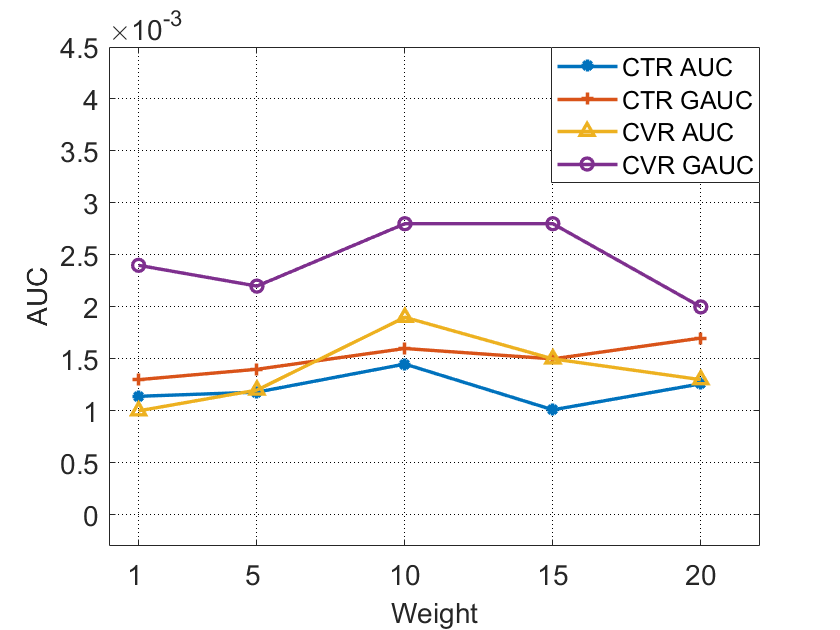}
  \label{fig_2_b}
  }
\centering
%\vspace{-0.3cm}
\caption{The impact of the parameters on the performance on Meituan recommendation dataset.}
\label{fig_param}
%\vspace{-0.3cm}
\end{figure}

\subsection{The Effect of Hyperparameters}
We explore how batch size and the weight $\lambda$ in Eq. \ref{newLoss} influence the performance, and show the results in Fig. \ref{fig_param}. For better demonstration, the records under each setting subtract the values of the baseline setting. Since the proportion of positive sample is much lower than that of negative sample, constructing the pair of positive and negative samples may not be accessible when the batch size is small. Most of the metrics get improved when the batch size increases from 64 to 384, and reach the peak at the batch size of 384, which means that 384 is a proper scope for constructing difficult pair of samples. The weight $\lambda$ has less effect compared to batch size, and the optimal performance is obtained by setting $\lambda=10$.

\section{CONCLUSION}
In this paper, we propose the PDAOM loss to directly optimize the AUC metric in the context of ranking models. By constructing difficult pair of opposite samples, we not only improve the performance of the binary classifier, but also reduce the computational complexity during training. 
We further explore how to incorporate the GAUC goal into the training objective.
Without loss of generality, we evaluate the DAOM loss in the tasks of CTR and CVR prediction of the recommendation system. Both offline and online A/B test demonstrate the effectiveness of the new objective function. We also discuss how the construction of the surrogate objective function and the parameters influence the performance. Besides, we prove that the proposed PDAOM loss works for both active users and inactive users.

\bibliographystyle{IEEEtran}
\bibliography{IEEEfull}

\end{document}